# Precise Hybrid-Actuation Robotic Fiber for Enhanced Cervical Disease Treatment

Jinshi Zhao[1†], Qindong Zheng[2†], Ali Anil Demircali[1], Xiaotong Guo[3], Daniel Simon[1,4], Maria Paraskevaidi[1], Nick W F Linton[2], Zoltan Takats[1,4], Maria Kyrgiou[1], Burak Temelkuran[1,4*]

[1]Department of Metabolism, Digestion and Reproduction, Faculty of Medicine, Imperial College London, London SW7 2AZ, UK.
[2]Department of Bioengineering, Faculty of Engineering, Imperial College London, London SW7 2AZ, UK.
[3]Department of Electrical and Electronic Engineering, Faculty of Engineering, Imperial College London, London SW7 2AZ, UK.
[4]The Rosalind Franklin Institute, Didcot OX11 0QS, UK
†These authors contributed equally to this work
*Corresponding author. Email: b.temelkuran@imperial.ac.uk

**Abstract**

Treatment for high-grade precancerous cervical lesions and early-stage cancers, mainly affecting women of reproductive age, often involves fertility-sparing treatment methods. Commonly used local treatments for cervical precancers have shown the risk of leaving a positive cancer margin and engendering subsequent complications according to the precision and depth of excision. An intra-operative device that allows the careful excision of the disease while conserving healthy cervical tissue would potentially enhance such treatment. In this study, we developed a polymer-based robotic fiber measuring 150 mm in length and 1.7 mm in diameter, fabricated using a highly scalable fiber drawing technique. This robotic fiber utilizes a hybrid actuation mechanism, combining electrothermal and tendon-driven actuation mechanisms, thus enabling a maximum motion range of 46 mm from its origin with a sub-100 μm motion precision. We also developed control algorithms for the actuation methods of this robotic fiber, including predefined path control and telemanipulation, enabling coarse positioning of the fiber tip to the target area followed by a precise scan. The combination of a surgical laser fiber with the robotic fiber allows for high-precision surgical ablation. Additionally, we conducted experiments using a cervical phantom that demonstrated the robotic fiber's ability to access and perform high-precision scans, highlighting its potential for cervical disease treatments and improvement of oncological outcomes.

Keywords: robotic fiber, hybrid-actuation, precision, cervical cancer

**Introduction**

Cervical cancer, the fourth most common cancer in women, caused about 311,000 deaths in 2018[1], with projections estimating this number will increase to 400,000 in 2030.[2] Early-stage cervical cancer (Stage 1A) and high-grade precancerous lesions can be treated with local excisional treatment approaches that remove the diseased part of the cervix while preserving fertility.[3] Common surgical approaches include cone biopsy, large loop excision of the transformation zone (LLETZ), cryotherapy, and trachelectomy (radical vaginal/abdominal trachelectomy (RVT/RAT)), which aim to completely remove cancer.[4,5] The operational precision of these in situ cancer resection or destruction techniques primarily depends on the surgeon's experience. Upon completion of these treatment procedures, a biopsy should be conducted to confirm the negative margins (healthy tissue surrounding the cancerous tissue), which indicates complete cancer removal.

However, based on a systematic review and meta-analysis by Arbyn et al., among 44,446 cervical precancer treatments, there is an overall 23.1% rate of incomplete excisions resulting in positive margins.[6] This incomplete excision increases the risk of residual or recurrent cancer, which may lead to additional treatments (such as radiation therapy, chemotherapy, or another surgery) and even reduce the long-term survival likelihood of patients. Beyond assessing surgical margin clearance, the expansion of depth and volume of cervical tissue removal or destruction is associated with an increased risk of preterm birth in later pregnancies.[7-10] Considering these complications from cervical disease treatments, surgical approaches should aim to completely remove the disease while minimizing the loss of healthy cervical tissue.[11]

The robotic surgery incorporated into the treatment of early-stage cervical cancer enables surgeons to enhance precision, ensure safety, and optimize surgical performance.[12, 13] The da Vinci robotic system, for instance, has been utilized in robotic radical trachelectomy (RRT) for early-stage cervical cancer, with its first application reported in 2008.[14] Subsequently, RRT has been performed in numerous cases, in which operation precisions were reported at the millimeter level.[13] For fertility-sparing RRT, yielding an overall live birth rate of approximately 50% for women undergoing this surgical procedure.[13] Additionally, the use of the da Vinci instrument increases the cost of the surgical procedure, which is a necessary factor to weigh against the substantial benefits.

Within the low-cost and compact robotic platforms, soft robots are less likely to cause trauma and pain,[15] which are preferred in surgeries near sensitive tissue. There is potential for addressing cervical cancer through various soft actuators via a transvaginal approach. Among soft manipulators, the tendon-driven catheter is one of the most frequently utilized in the medical field, which presents high flexibility and force output.[16-18] Leber et al. have presented various designs of fiber-based tendon-driven actuators that are fabricated using thermal drawing techniques.[19] However, the performance of tendon-driven instruments remains limited by non-linear friction problems, such as hysteresis, dead zone, and plastic torsion,[20] which requires complex control models to compensate.[21,22]

Electrothermal actuators are actuated using the thermal expansion behavior of solids. The inherent characteristics of this type of actuators are small range and high precision motion, making them usually used for microelectromechanical systems (MEMS) as a micro-positioner or a micro-grasper.[23-26] In previous work, we successfully developed and navigated an electrothermal-actuated fiber-based robot, fabricated using the fiber thermal drawing technique.[27] This technology involves embedding resistance wires into a polymer fiber during the fiber drawing process. When applying voltages to these wires, the Joule heat generated from the wires increases the temperature of the surrounding polymer, causing a temperature gradient across the fiber's cross-section. This gradient induces non-uniform thermal expansion within the polymer, subsequently causing deflection in the fiber. By programming the voltage applied to the resistance wires, the fiber motion can be controlled with high precision.[27]

This approach demonstrates the advantages of fiber robots in the following perspectives. (i) Small scale: the 1.7 mm diameter fiber allows access to challenging surgical sites without blocking the view of the surgeon. (ii) High motion precision: the motion precision of a 10cm-long fiber is below 50 micrometers. (iii) Availability for mass production: hundreds of meters long low-cost polymeric fiber can be produced from each time fabrication. (iv) Integration with functional tools: introducing the surgical laser fiber with the robotic fiber enables precise surgical ablation. However, this approach is limited in its range of motion within 5x5 mm$^2$ (depending on the fiber's length), requiring manual adjustment of the robot's position when the target ablation area is out of reach.[27]

This work presents the design and development of a fiber-based robot for transvaginal laser surgeries, as schematic illustrated in Figure 1A. This robotic fiber, with a single continuum design, is driven by both tendon-driven and electrothermal actuation mechanisms (Fig. 1B), enabling sub-100μm precision and a 4.6 cm motion range from its origin. Tendon-driven actuation allows flexible motion, while electrothermal actuation enables high-precision motion in a small range (4x4 mm$^2$). We outlined the fabrication process of the fiber body, which includes a highly scalable fiber drawing technique. Subsequent sections demonstrated the kinematics, simulations, and characterization results of this robotic fiber. For robot motion control, telemanipulation is employed for the tendon-driven actuation, and pre-defined trajectory control for the electrothermal actuation. Additionally, we combined robotic fiber with carbon dioxide ($CO_2$) laser fiber, enabling precise surgical scan and ablation. We validate the prototype's functionality through its scanning of a simulated cervical cancer model in a gynecological phantom, serving as a proof-of-concept of a cervical disease treatment device.

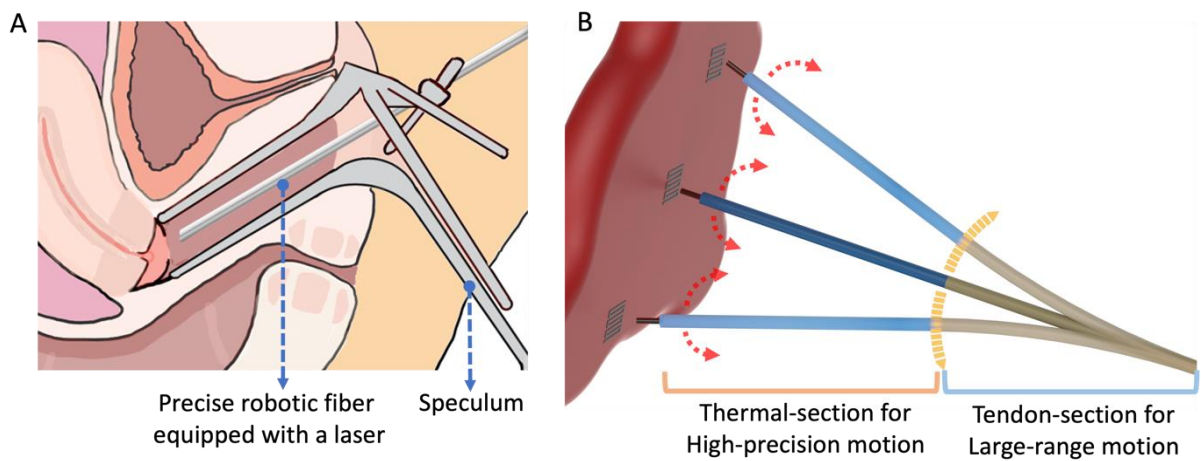

**FIG. 1.** Proposed approach for early-stage cervical cancer surgery. (**A**) A robotic fiber delivers laser fiber to perform precise transvaginal scans atop the cervix. (**B**) Schematic representation of the robotic fiber with thermal (blue) and tendon (brown) sections.

**Materials and Methods**

To accommodate the majority of transvaginal surgeries, we propose a miniaturized fiber-based robot that is 150 mm long and 1.7 mm in diameter. This robotic fiber is capable of achieving multi-directional motion through a hybrid-actuation mechanism (Fig. 1B). The first 70 mm from the distal end of the robotic fiber utilizes the electrothermal actuation mechanism, referred to as the 'thermal section' throughout the remainder of the text. The remaining section of the fiber utilizes the tendon-driven mechanism, referred to as the 'tendon section'.

*Robotic fiber design and fabrication*
The robotic fiber consists of three main components: (1) a polymeric fiber body with a custom-designed multi-lumen cross-section to house functional wires for robot actuation, (2) resistance wires that serve as the heating source for electrothermal actuation, and (3) pulling wires for tendon-driven actuation. The subsequent sections provide detailed fabrication methodologies.

The polymeric fiber body was fabricated using fiber drawing technology. We began with the fabrication of a fiber preform, which was manufactured by drilling holes according to the desired cross-sectional design (Fig. S1) in a thermal plastic rod (polycarbonate (PC), 400 mm outer diameter, 150 mm long). The prepared polymer rod was then placed into a vacuum oven (at 70 °C) to degas the polymer for seven days. Subsequently, the preform was initially placed into the heating furnace of a fiber drawing tower. This furnace is a vertical cylindrical design, featuring three separate temperature zones, set at 150 °C, 250 °C, and 85 °C respectively, as illustrated in Fig. 2A. The primary goal of this stage is to heat-soften the fiber preform, enabling the preform scales down into the fiber while preserving the shape of its cross-section. Throughout the drawing phase, the fiber preform was consistently down-fed into the heating furnace at a steady rate of 1 mm/min. Simultaneously, a motorized capstan exerted a pulling force on the preform's lower end. This capstan operated at a controllable line speed of approximately 550 mm/s, offering a stable fiber diameter of 1.7 mm (Fig. 2A).[27]

This drawn fiber was then cut into several 150 mm long sections (length adjustable on demand) to serve as the polymeric fiber body for the robot. This specific length of the fiber was selected in accordance with the typical vaginal length, which ranges from approximately 68 to 150 mm.[28] The diameter of the fiber was designed to be minimized while sufficiently accommodating a surgical $CO_2$ laser fiber (0.9 mm in diameter) in a center channel and the actuation wires in side lumens. Fig. 2B shows the cross-sectional view of the fiber, which equips three pairs of channels for heating wires allocation and three equilateral-triangularly distributed channels for tendon wires allocation.

The heating wires were embedded into the polymer fiber by post-feeding 50 μm diameter molybdenum wires (electrical conductivity at 20 °C: $18.7 \times 10^6$ S/m) from the pairs of channels at the distal end of the fiber and forming U-shapes. The fed wires were then exposed out of the fiber from pre-cut holes located 70 mm away from the distal end, forming a 70 mm long electrothermal actuated section (Fig. 2C). Six exposed molybdenum wires were soldered with 50 μm diameter copper wires (electrical conductivity at 20 °C: $59.6 \times 10^6$ S/m) that possess a significantly lower resistance. These copper wires would be distributed with a smaller voltage than the molybdenum wires when power is supplied, thus allowing the molybdenum wires to act as a heating source and creating a 70 mm long selective actuation section. A heat shrink tubing was used to cover the top of the soldering point to isolate the electricity. The copper wires were then wrapped around the rest of the fiber to prevent the heating wires from being pulled due to the bending motion of the tendon section. The remaining parts of the copper wires were soldered onto the specific pins of a circuit adapter at the proximal end of the fiber, which was connected to a detachable actuation circuit (Fig. 2C).

The tendons are composed of three molybdenum wires (80 μm in diameter). Each tendon was inserted into the tendon channel from the pre-cut hole 80 mm away from the distal end. Following that, this tendon was exposed out of the fiber from another pre-cut hole that 150 mm away from the distal end. The tendon was fixed at the 80 mm position by wrapping it with stainless-steel wire and dipping it with glue. The section with tendons embedded forms the tendon section.

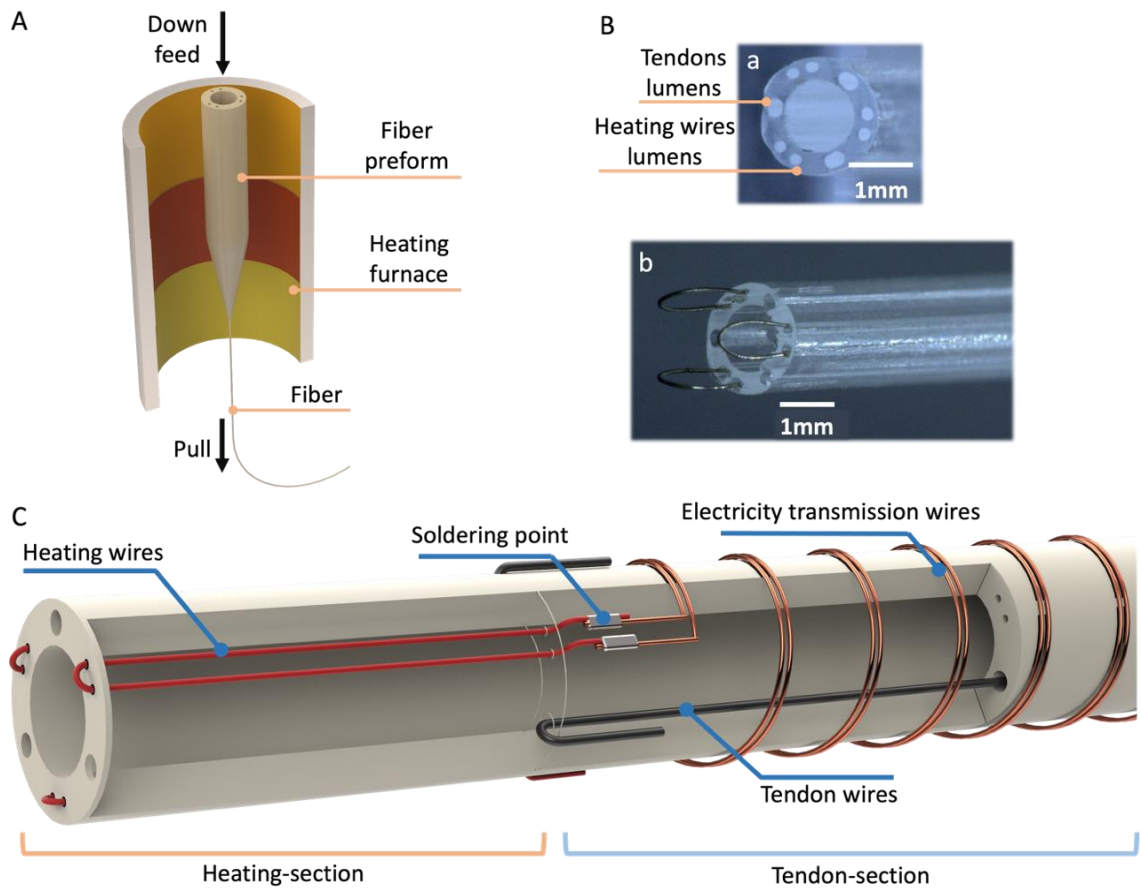

**FIG. 2.** Design and fabrication of robotic fiber. (**A**) Illustration of the fiber drawing technique. (**B**) Microscopic views of (a) the fiber's cross-section and (b) the fiber with embedded resistance heating wires. (**C**) Illustration detailing the mechanical structure of robotic fiber.

*Actuation unit and control system*
To control two sections of the fiber robot, two actuation units were employed: a parallel motorized-leadscrews unit for manipulating the tendon section and a voltage actuation unit for manipulating the thermal section (Fig. S2). For the tendon section, three tendons that came out from pre-cut holes on the fiber were fixed with bespoke nut blockers, allowing for symmetric wire pulling during the linear translation of the tendon-driving unit (Fig. 2C). This tendon-driving unit (Fig. 2A) consists of a triple parallel motorized-leadscrew linear actuator (Fig. S3A and B), of which the motion was controlled by an open-loop control algorism built in LabVIEW (version 2018, National Instruments, USA). (The details of actuation system are outlined in Note S1, Supplementary).

The tendon-driving unit controller was developed based on our kinematic model, which used the constant-curvature (CC) approximation for the tendon section (80 – 150 mm section from the fiber's distal tip). This kinematic model was inspired by the kinematic model developed by Leber et al[16]. The thermal section was assumed to be a straight line that tangent to the tendon section (Fig. S4A). Based on these two assumptions, we used inverse kinematics to obtain the relationship between the displacement of each pulling wire and the two-dimensional coordinates of the fiber's distal tip. The detailed kinematic model for tendon section control is described in Supplementary Note S2 and Figure S3.

The thermal section's actuation system is illustrated in Figure S2. Control instructions from the laptop are transmitted to a real-time controller (cRIO-9025, National Instruments, USA). This controller links to a digital-to-analog converter (NI-9264 DAC, National Instruments, USA), which transmits analog signals to a custom-designed actuation circuit. The actuation circuit was crafted as printed circuit boards (PCB) with amplifiers to provide power to drive the fiber's actuation.[27] A power cable was deployed to link the PCB and the resistance wires of the robotic fiber, enabling electricity transmission. The thermal section control principle is based on the displacement of the fiber that is proportional to the magnitude of the electric power applied to the corresponding wires.

$$\text{Equation: } D = \alpha \cdot P = \alpha \cdot \frac{V^2}{R}$$

Where D is the displacement of the fiber distal tip, P is the applied power, V is the applied voltage, and α is a constant derived from the ratio between power and displacement characterization.

The bending direction and the displacement of the thermal section can be calculated from the sum of the power vectors in the corresponding directions.[27] The details of the thermal-section control algorithm's principle are described in Note S3.

*This section includes the methodologies and results of the characterization, simulation, surgical tools integration, as well as phantom testing of the fiber robot.*
**Results**

*Finite element analysis (FEA) for thermal section*
To demonstrate the concept and predict the performance of thermal section motion, we employed finite element analysis (FEA) within the COMSOL Multiphysics® 5.6 (COMSOL AB, Sweden). Detailed simulation procedures, mesh methods, material parameters, and ambient settings are described in Supplementary Note 3. In the simulation, we applied electrical power ranging from 0 to 1.2 W (in intervals of 0.1 W) to one side of the resistance wires. Following this, the temperature distribution and the structural deflection of the robotic fiber in the steady state were recorded (Fig. 3A-D). The outcomes of the simulation data described a highly linear relationship between the applied actuation power and both the highest fiber surface temperature and the maximum surface temperature difference (Fig. 3B). Additionally, the results also present a linear relationship between applied power and the displacement of the fiber's distal tip (Fig. 3D).

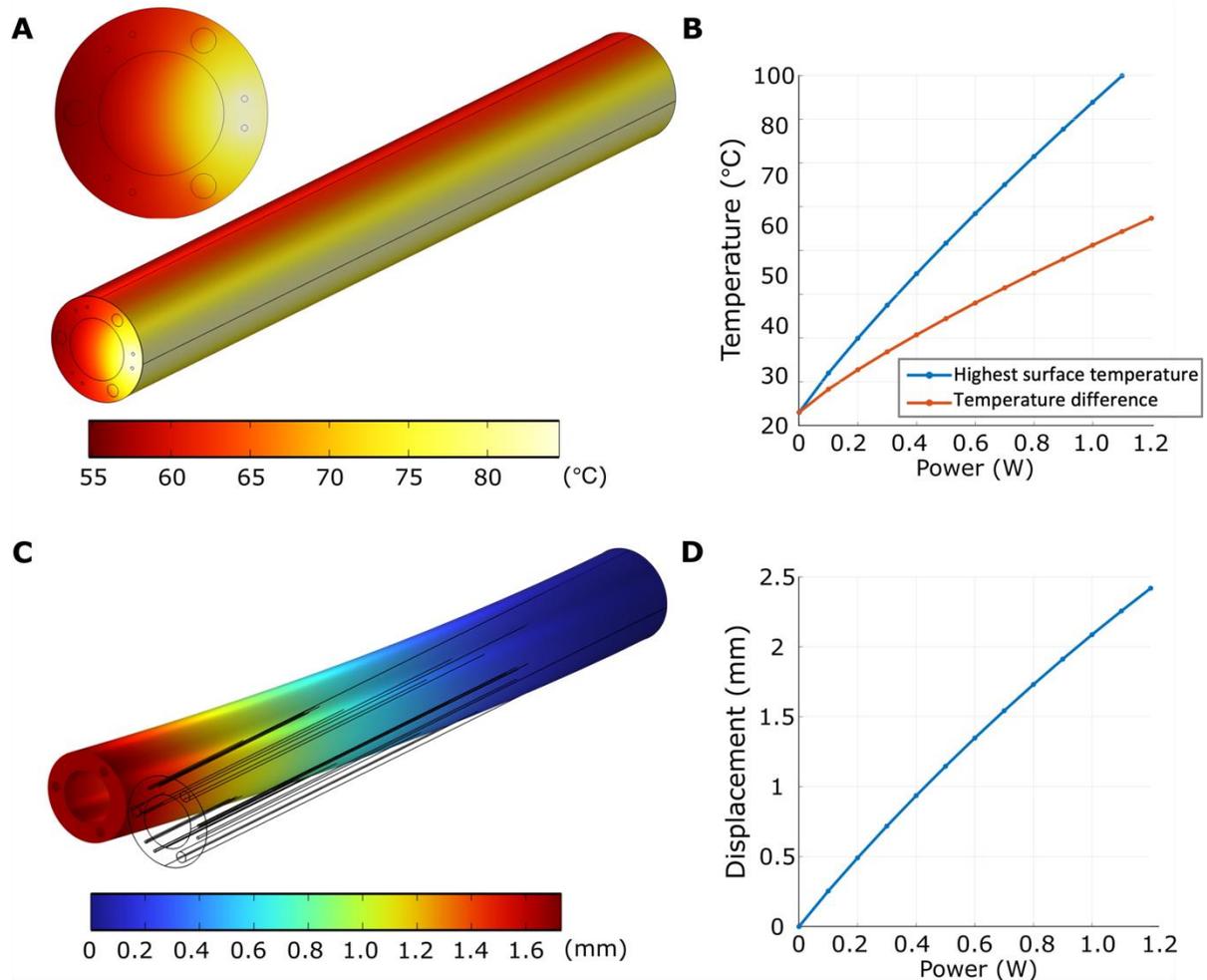

**FIG. 3.** Simulation results of the thermal section. (**A**) Simulation of the robotic fiber's temperature distribution under 0.8 W applied power. (**B**) Plot of the highest surface temperature and the maximum surface temperature difference within the fiber under various applied powers. (**C**) Simulation of robotic fiber's structural deflection under 0.8 W input power. (**D**) Plot of the fiber tip displacement under various applied powers.

*Characterization of the robotic fiber*
The maneuverability characterization of the robotic fiber includes the performance evaluations of both the tendon and thermal sections. During the tendon section characterization, the fiber's thermal section remained in its resting position and inactivated. By pulling one of the tendons from 0 mm to 0.9 mm with a step size of 0.1 mm, the tendon section was bent in a single direction with varying angles. A camera was positioned above the robotic fiber to capture the deflection during manipulation, from which, the view of overlapped fiber displacement due to tendon pulling is shown in Figure 4A. Figure 4B presents the measurement results, illustrating a linear relationship of the robotic fiber tip displacement (within the range of 46 mm) in response to each step of tendon pulling length. A minor tendon pulling of 0.1 mm induced approximately a 50-fold displacement (5 mm) at the distal end of the fiber robot.

The maneuverability analysis of the fiber's thermal section was performed on an independent thermal section of the robotic fiber. To precisely track the fiber robot's position, an optical fiber (FP200URT, Thorlabs Inc., USA) with a 200 µm diameter was inserted into the thermally actuated fiber's central channel. This optical fiber was connected to a 650 nm wavelength photodiode at its proximal end, allowing for a visible red dot at its (and robotic fiber's) distal end. Subsequently, a high-resolution camera (VHS-2000, Keyence Co., Japan) was mounted in front of the robotic fiber's distal end to track the motion of the fiber tip (Fig. 4C).[27] During the thermal section characterizations, power levels ranging from 0 to 1.2 W (in a step increment of 0.1 W) were applied to one pair of heating wires. Each power level was applied for 100 seconds. The result in fiber tip displacement was recorded in Figure 4D. A linear relationship between the fiber tip's displacement and the applied power was observed when the power was below 0.8 W, within a displacement of 1.95 mm. Also, the steady-state error during this period was below 100μm. However, when the power exceeded 0.8 W, the fiber robot's displacement became non-linear. This nonlinearity is likely due to excessive heat from the resistance wires, causing the polymer to melt and potentially leading to irreversible deformation. Thus, the thermal section should operate with power below 0.8 W, where the motion range is below 2 mm from the origin.

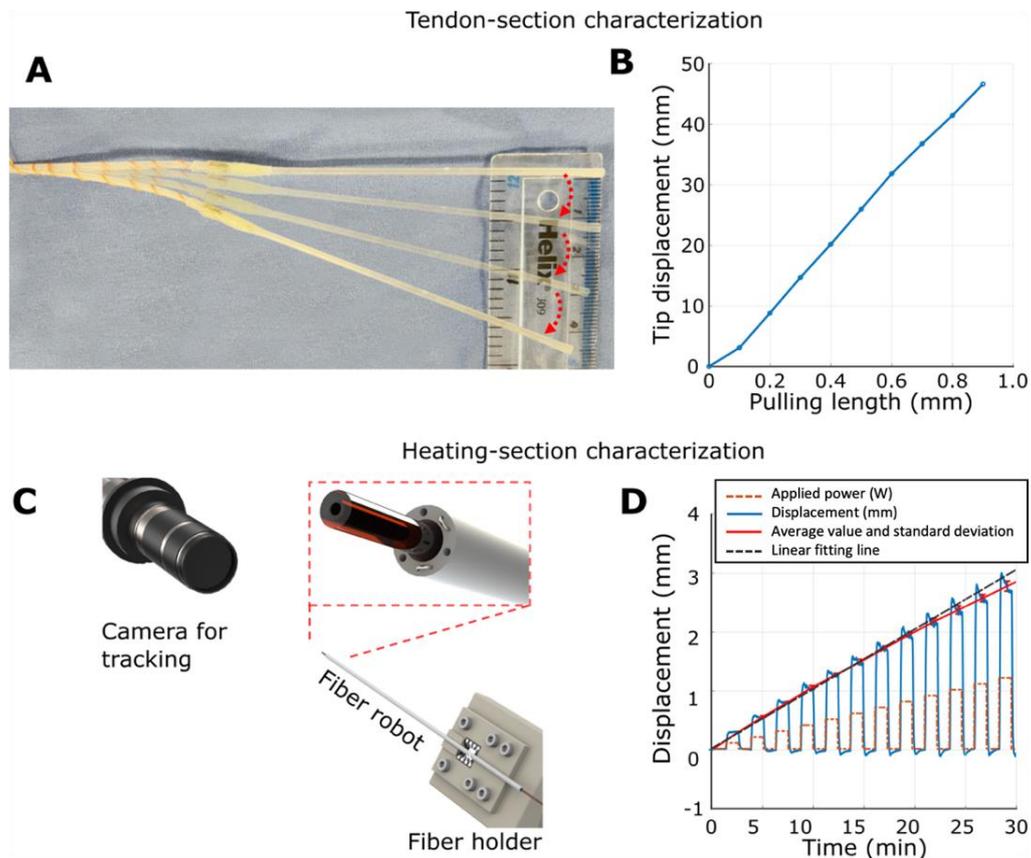

**FIG. 4.** Fiber robot maneuverability characterization. (**A**) Overlay of robotic fiber displacement when pulling tendons with step increment length. (**B**) Plot of the relationship between the tendon pulling length and the fiber tip displacement. (**C**) Schematics of the fiber tip tracking setup for the thermal section characterization. (**D**) Plot of the relation between the applied power to heating wires and the displacement of the electrothermally actuated fiber tip. The red line connects each displacement's average value, while the straight black dashed line indicates the desired positions.

*Robotic fiber with surgical laser*
To functionalize the robotic fiber in medical applications, we integrated a $CO_2$ laser fiber (0.9 mm dia., OmniGuide, USA) with the robotic fiber by inserting it through the robot's central working channel (Fig. 5A). This $CO_2$ laser fiber allows transmission of 10.6 μm wavelength laser from a laser emitter (FELS-25A Intelliguide, OmniGuide Inc., USA) to the target tissue for ablation. Upon actuation and control of our robotic fiber, the laser fiber can be navigated to the target place, enabling the ablation and precise scan of tissues.

The maneuverability of the robotic fiber with laser fiber was evaluated via an ex-vivo laser ablation experiment on bovine tissue. In this experiment, the tissue employed exhibited a distinct net pattern of adipose tissue, visually distinguished by its white appearance compared with adjacent muscle tissues. The experiment was designed to ablate one primary strand of adipose tissue using the robotic system.

During the experiment, the telemanipulation control method was employed to control the movement of the tendon section, guiding the tip of the robot fiber to the top-central region of the desired adipose tissue. The laser was inactive during this phase. Once the tip of the fiber robot stabled at the desired position, the thermal section was controlled to move along a predefined 'Raster' path, enabling comprehensive scanning with a rectangular coverage. The laser was switched to active status during this phase. Subsequently, the tendon section was controlled to guide the fiber tip to the next place along the desired strand of adipose tissue, followed by the next raster path scanning. These processes were repeated until the targeted adipose tissue strand was fully covered, on which the ablation area can be visually distinguished by its burned markers (Fig. 5C). At the end of ablations, the fiber robot was controlled to implement a swing-back motion to ablate the gap left by previous scans (Fig. 5C (e, f)), which further demonstrates the maneuverability of the system. Further details are demonstrated in Supplementary Video S1. From the results, the robotic system delivered the laser fiber to implement a full coverage and precise ablation of the targeted adipose tissue strand, with minimal damage to the surrounding muscle tissues.

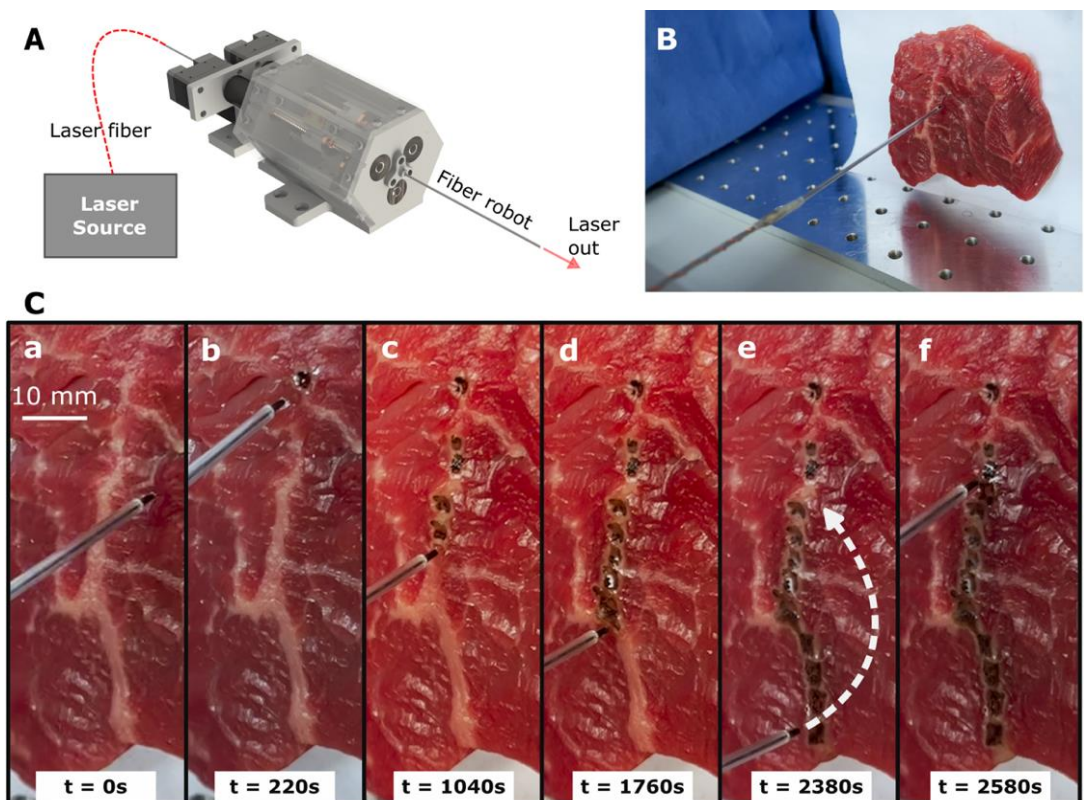

**FIG. 5.** Laser fiber integration and ex-vivo ablation experiment. (**A**) Schematics of laser fiber assembly setup. (**B**) The experimental setup for the ex-vivo bovine tissue ablation experiment. (**C**) Ablation processes with time markers.

*Gynecological Phantom study*

To evaluate the clinical performance of the robotic system in transvaginal surgical procedures, we conducted an experiment using a gynecological phantom. The experimental setup is depicted in Figure 6A. The gynecological phantom, comprising an external genital, vagina, and cervix with early-stage cancer (Fig. 6B (a)), was positioned on a lab bench to mimic a patient's posture during a colposcopy test. The robotic fiber, with its actuation unit, was mounted on an XYZ linear translation stage (LT3, Thorlabs Inc., USA). This stage aided in the axial insertion of the robotic fiber into the vaginal area. An optical fiber connected to a 650 nm wavelength photodiode, simulating the laser fiber and its source, was coupled with the robotic fiber. The red spotlight emitted from the fiber's tip served as a simulated ablation point (Fig. 6B (b)). An endoscopic camera was configured at a position with a clear view of the fiber tip and cervix, offering real-time visualization during the study. Additionally, a thermal camera (A400, FLIR Systems Inc., USA) was employed to monitor and record the temperature distribution along the robotic fiber (Fig. 6A).

The demonstration aimed to guide the robotic fiber to scan exclusively over the simulated cervical cancer area (the red region in Figure 6B (a)). Initially, the fiber was introduced into the vaginal orifice using the translational stage. Upon reaching the desired depth with the robotic fiber's tip (approximately two millimeters from the cervix's surface), its tendon section was telemanipulated to the top of the targeted tumor (Fig. 6C). After resting the tendon section's movement, the laser diode was activated, generating a red dot on the fiber tip, signifying the commencement of laser ablation. The thermal section subsequently drove the simulated laser fiber to perform a precise raster scanning. Once the raster scanning was completed, the simulated laser was deactivated. Following the same protocol, we telemanipulated the tendon section to shift the fiber tip to the next position. This procedure was repeated three times, ensuring the majority of the simulated tumor was scanned.

To validate the scanning results, the optical fiber tip's position underwent post-tracking using MATLAB (version 2019, MathWorks Inc., USA) for RGB color tracking of the red dot on the fiber tip. The three raster scans were differentiated by blue, red, and yellow markers, respectively, as displayed in Figure 6C. However, due to the absence of burn marks from real laser ablation on the phantom, the scanned path was invisible during the experiment, resulting in certain tumor areas remaining unscanned. This problem can be rectified in the real surgical scenario, where surgeons can refine scanning positions based on the ablation burn marks, as evidenced in the bovine tissue ablation (Fig. 5). In addition, during the robotic fiber's operation, we monitored the fiber's temperature distribution. The thermal section's peak temperature reached 75 °C, with the tendon section registering a minimal temperature rise (Fig. 6 D and E). Supplementary Video S2 provides further details of this study.

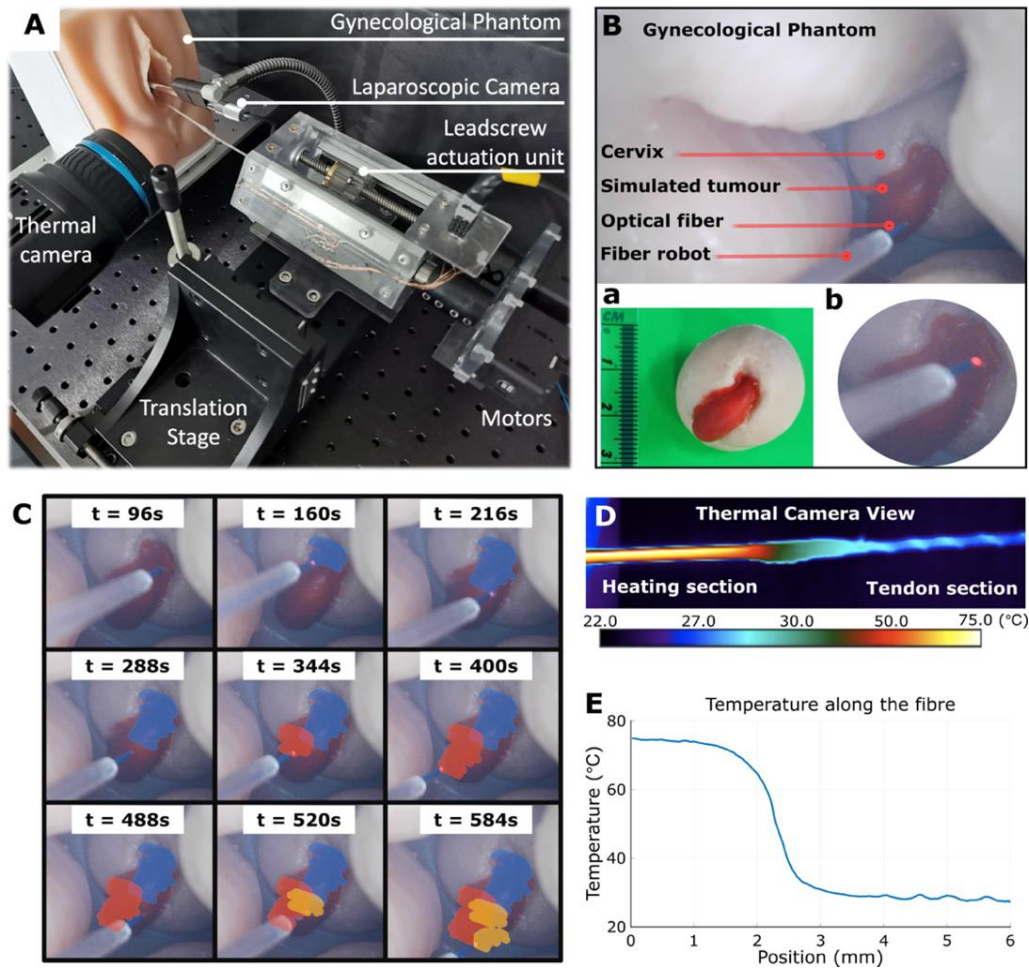

**FIG. 6.** Gynecological phantom study. (**A**) Phantom study experimental setup. (**B**) The endoscope view inside the phantom shows the tumor (a) and the fiber tip (b). (**C**) Simulated tumor scanning procedures in time sequence. Each time of scan is highlighted in blue, red, and yellow in sequence. (**D**) Thermal camera view of the actuated robotic fiber, showcasing temperature distribution along the fiber. (**E**) Maximum temperature along the robotic fiber.

## Conclusions

We have developed a 150 mm long and 1.7 mm diameter robotic fiber, employing a hybrid actuation system including thermal and tendon-driven actuation mechanisms. This approach allows the utilization of low-cost materials and low-profile fabrication methods to develop robotic surgical instruments. The development of control algorithms allows the telemanipulation of the tendon section and predefined path control of the thermal section, thereby facilitating the strategical manipulation of surgical laser fibers. We conducted motion characterization for this robot to evaluate its performance, from which the robotic fiber facilitates a 46 mm motion range from its origin and a sub-100 µm scanning precision.

This work is motivated by the lack of precise intraoperative surgical instruments for enhancing the clearance of surgical margins and fertility-sparing surgical treatments. The design of this robotic fiber focuses on both motion precision and accessibility aspects, which potentially enhances the complete disease removal and minimizes the sacrificed healthy tissue. The integration of a $CO_2$ laser fiber functionalizes the robotic fiber for precise laser ablation surgeries. Additionally, we conducted a gynecological phantom study as the proof-of-the-concept for clinical applications, presenting scans on the simulated tumor area.

Follow-up work will focus on developing an autonomous scanning algorithm to ensure comprehensive coverage of suspected tissue regions. To achieve this, a visual servoing system will be developed for the robotic system to aid in visual diagnostics, path planning, and closed-loop control. Furthermore, we will combine the robotic fiber with molecular diagnostic sensing systems such as Rapid Evaporative Ionization Mass Spectrometry (REIMS) to build an intraoperative diagnostic and therapeutic device for cervical cancer. REIMS, a mass spectrometry-based real-time tissue classification system, analyzes the mass-to-charge ratio of aerosol resulting from surgical laser ablation to provide tissue type. Previous studies have validated its diagnostic capabilities in the intraoperative diagnostics of cervical diseases[29] and have incorporated with laser for identifying cervical cancers.[30] By combining the robotic fiber position information with the tissue identification information, our robotic system will be able to generate a diagnostic tissue map. This method would be more efficient than the intraoperative frozen biopsies used for assessing margin clearance[31], as it could eliminate time-consuming histopathology procedures and minimize harm to tissues.

**Author's Contributions**
J.Z., Q.Z. and B.T. proposed the ideal and conceived the project. J.Z., Q.Z., Z.T., M.K. and B.T. participated in the concept design. J.Z. and Q.Z. designed and fabricated the robot. J.Z., Q.Z. and D.S. conducted the experiments and data interpretation. A.A.D. performed the simulation. J.Z. and X.G. developed the mathematical model. J.Z., Q.Z., A.A.D. and X.G. performed the visualization. J.Z. and Q.Z. drafted the original article. D.S., M.P., N.W.F.L. and B.T. reviewed and edited the article. M.P., N.W.F.L., Z.T., M.K. and B.T. supervised the research. J.Z. and Q.Z. contributed equally to this work. All authors provided input in the discussion of the project.

**Author Disclosure Statement**
No competing financial interests exist.

**Supplementary Material**
Supplementary Note S1
Supplementary Note S2
Supplementary Note S3
Supplementary Note S4
Supplementary Figure S1
Supplementary Figure S2
Supplementary Figure S3
Supplementary Figure S4
Supplementary Figure S5
Supplementary Video S1
Supplementary Video S2